\documentclass{article}

\usepackage{arxiv}

\usepackage[utf8]{inputenc} 
\usepackage[T1]{fontenc}    
\usepackage{hyperref}       
\usepackage{url}            
\usepackage{booktabs}       
\usepackage{amsfonts}       
\usepackage{nicefrac}       
\usepackage{microtype}      
\usepackage{lipsum}
\usepackage{multirow}
\usepackage{orcidlink}
\usepackage{graphicx}
\graphicspath{ {./images/} }

\usepackage{threeparttable}
\usepackage{amsmath}

\title{\textsc{Neurai-VN} Benchmark: Standardized Machine Learning Models for Multimodal Digital Phenotyping in Mental Health Classification
}

\author{
 Quoc-Cuong Pham\,\orcidlink{0009-0005-6633-8201} \\
  College of Engineering and Computer Science, \\
  VinUni-Illinois Smart Health Center, \\
  VinUniversity, \\
  Hanoi, Vietnam. \\
  \texttt{24cuong.pq@vinuni.edu} \\
   \And
 Hoang-Thuy-Duong Vu\,\orcidlink{0009-0005-6775-6691} \\
  College of Engineering and Computer Science, \\
  VinUni-Illinois Smart Health Center, \\
  VinUniversity, \\
  Hanoi, Vietnam. \\
  \texttt{26duong.vht@vinuni.edu} \\
\And
 Thi-Thanh-Huong Ha\,\orcidlink{0000-0002-8884-8692} \\
 School of Biomedical Engineering, \\
 International University, \\
 Vietnam National University HCMC \\
 Ho Chi Minh, Vietnam. \\
  \texttt{htthuong@hcmiu.edu.vn} \\
\And
 Huy-Hieu Pham\,\orcidlink{0000-0003-4851-2518} \\
 College of Engineering and Computer Science, \\
 VinUni-Illinois Smart Health Center, \\
 VinUniversity \\
 Hanoi, Vietnam. \\
  \texttt{hieu.ph@vinuni.edu.vn} \\
}

\begin{document}
\maketitle
\begin{abstract}
Digital phenotyping (DP) using smartphones and wearable devices has emerged as a promising approach for assessing mental health, particularly depression and anxiety. 
However, progress remains difficult to evaluate because of heterogeneity across datasets and inconsistencies in preprocessing pipelines.
In this work, we introduce a reproducible machine learning benchmark using the \textsc{Neurai-VN} dataset, a multimodal digital phenotyping dataset collected from 100 Vietnamese adults over two weeks. 
We define four binary classification tasks evaluated using standardized subject-wise cross-validation. Representative linear, tree-based, and neural baseline models are evaluated systematically across predefined feature-group configurations.
Mean subject-level F1 scores across five cross-validation folds reached 0.71 for Healthy Control vs. Depression and Healthy Control vs. Clinical, while Healthy Control vs. Anxiety and Depression vs. Anxiety achieved 0.69 and 0.56, respectively.
These baseline results provide reproducible baselines for future research on multimodal DP for mental health classification tasks.
The code to reproduce the benchmark is available at \url{https://github.com/neurai-vn/Neurai-VN-benchmark}.

\end{abstract}


\section{Introduction}
\label{sec:intro}
Digital phenotyping (DP) transforms heterogeneous sensing data, collected across individuals and time, into behavioral representations that can be mapped onto clinical outcomes such as depression and anxiety \cite{akre2024advancing}.
%
Smartphones and wearable devices make this feasible at scale: passive streams such as mobility dynamics, screen interaction, sleep, and cardiac activity, together with active self-report prompts, capture day-to-day variation that periodic clinical assessment is poorly positioned to observe, given that it is administered weeks apart and relies on retrospective recall \cite{insel2018digital, mohr2017personal}.
The clinical premise is therefore straightforward: if routinely collected behavioral signals track symptom severity, they can support earlier detection and continuous monitoring at negligible marginal cost to the patient.


A decade of empirical work has established that such associations exist. Early studies linked GPS-derived mobility features to depressive symptom severity \cite{saeb2015mobile, canzian2015trajectories}, and longitudinal campus deployments related passive sensing to mental health and academic outcomes \cite{wang2014studentlife, chikersal2021detecting}.
%
More recent studies report similar findings across a widening range of signals: depressive symptoms have been associated with screen time \cite{pieh2025smartphone, li2025cross}, phone unlocks \cite{zhang2025nomophobia}, and social media exhaustion \cite{guo2026linking}, while anxiety disorders have been consistently linked to reduced heart rate variability (HRV) \cite{fagioli2025heart} and abnormal sleep durations \cite{riemann2025chronic}.
Several public resources have supported this line of work, including StudentLife \cite{wang2014studentlife}, CrossCheck \cite{ben2017crosscheck}, Tesserae \cite{mattingly2019tesserae}, LifeSnaps \cite{yfantidou2022lifesnaps}, and GLOBEM \cite{xu2023globem}.
Cross-study comparison nevertheless remains difficult, for three distinct reasons. First, datasets differ in population, sensing modality, and sampling regime, so a model reported on one cohort is rarely applicable to another without redesign. Second, feature extraction and preprocessing are typically task-specific and only partially documented, which limits reproducibility \cite{mcdermott2021reproducibility}. Third, and most consequentially, evaluation protocols are not aligned: record-wise cross-validation, which allows observations from the same participant to appear in both training and test folds, yields substantially optimistic estimates relative to subject-wise validation \cite{saeb2017need}, and studies further differ in how continuous symptom scales are binarized into class labels.
The practical result is that reported performance figures are not comparable across studies, and that models generalize considerably less well across cohorts and collection periods than single-dataset results suggest \cite{xu2023globem}.
A further limitation is demographic: existing resources are drawn overwhelmingly from Western, English-speaking, and predominantly university-affiliated samples \cite{henrich2010weirdest}, leaving it unresolved whether reported behavior–symptom associations hold in other linguistic, cultural, and socioeconomic settings.

In this work, we present a reproducible benchmark for mental health classification tasks on the \textsc{Neurai-VN} \cite{cuong_q_pham_2026_21969065}, a multimodal digital phenotyping dataset using real-world Vietnamese cohort. 
Our objectives are to: (1) define clinically relevant modeling tasks; (2) standardize feature configurations; (3) establish baseline models; and (4) adopt a unified subject-wise cross-validation protocol for fair and reproducible evaluation.
The resulting baseline performance establishes reference points for future method development and enables consistent comparison across models.


\section{Dataset}
\label{sec:method}
We used data from \textsc{Neurai-VN}~\cite{cuong_q_pham_2026_21969065}, a mental health cohort from 100 Vietnamese adults over a two-week period under realistic condition.
The dataset includes 13 passive-sensing modalities collected from wearable and smartphone devices at multiple temporal scales. These include wearable signals (e.g., activity, heart rate, sleep, and SpO2), smartphone signals (e.g., accelerometer, gyroscope, and app state), and active self-reports, with demographic and clinical outcome labels provided in the metadata.

Table~\ref{tab:dataset_characteristics} summarizes the characteristics of the cohort. 
Participants were labeled as depressive (\textit{Dep}), anxiety (\textit{Anx}), healthy controls (\textit{HC}), or other psychiatric conditions (\textit{OPC}), excluding depressive and anxiety disorders.

\begin{table}[h]
\centering
\small
\begin{threeparttable}

\caption{Characteristics of the \textsc{Neurai-VN} cohort.}
\label{tab:dataset_characteristics}

\setlength{\tabcolsep}{8pt}
\renewcommand{\arraystretch}{1.05}

\begin{tabular}{p{8cm}p{3cm}}
\toprule
\textbf{Characteristic} & \textbf{Value} \\
\midrule

\multicolumn{2}{l}{\textit{Study Group}} \\

Healthy controls (\textit{HC}) & 43 \\
Individual with Depressive Depression (\textit{Dep}) & 31 \\
Individual with Anxiety Disorder (\textit{Anx}) & 18 \\
Other psychiatric conditions (\textit{OPC}) & 8 \\
Total participants & 100 \\

\addlinespace
\midrule
\multicolumn{2}{l}{\textit{Demographics}} \\

Female, $n$ (\%) & 65 \\
Age (years), mean $\pm$ SD & 26.1 $\pm$ 4.6 \\
Age range & 19--42 \\

\bottomrule
\end{tabular}


\end{threeparttable}
\end{table}

\newpage
\section{Method}
\label{sec:method}
This section describes the methodology for establishing baseline machine
learning performance on the \textsc{Neurai-VN} dataset. We introduce the task
definition, data preprocessing, feature representation, and benchmark design
to provide a standardized and reproducible evaluation framework.
\subsection{Data Representation}
\textbf{Modality Grouping. }
The released dataset comprises raw recorded data modalities, organized into six modality groups based on sensing source and temporal resolution (Table~\ref{tab:modality_groups}).

\begin{table}[!h]
\centering
\small
\begin{threeparttable}

\caption{Feature groups evaluated in the benchmark experiments.}
\label{tab:modality_groups}

\setlength{\tabcolsep}{6pt}
\renewcommand{\arraystretch}{1.0}

\begin{tabular*}{\textwidth}{@{\extracolsep{\fill}}lllp{7.4cm}}

\toprule
\textbf{Modality Group} &
\textbf{Data Source} &
\textbf{Temporal Resolution} &
\textbf{Included Data Records} \\
\midrule

$W_m$
&
Wearable
&
Per-minute
&
\texttt{ats}, \texttt{hrts}, \texttt{azmts}
\\

$W_s$
&
Wearable
&
Daily
&
\texttt{sleep}, \texttt{hrv}, \texttt{spo2},
\texttt{br}, \texttt{skinTemp}
\\

$S_d$
&
Self-report
&
Daily
&
\texttt{dailySurvey}, \texttt{moodLog}
\\

$S_w$
&
Self-report
&
Weekly
&
\texttt{phq9}, \texttt{gad7}
\\

$P_e$
&
Smartphone
&
Event-based
&
\texttt{appstate}
\\

$P_c$
&
Smartphone
&
Continuous
&
\texttt{accelerometer},
\texttt{gyroscope}
\texttt{network},
\texttt{battery}
\\

\bottomrule

\end{tabular*}


\end{threeparttable}
\end{table}
\noindent
\textbf{Modality Formulation. }
Given a cohort of $N$ subjects, we denote the $i$-th subject as $s_i$, where $i\in\{1,\dots,N\}$. Each subject is observed over $K_i$ days, and each observation is indexed by a subject-day pair $(i,k)$, where $k\in\{1,\dots,K_i\}$. Each subject is associated with a label $y_i$, representing the outcome for the prediction task.

Each subject-day pair $(i,k)$ is represented by two complementary data types: a collection of modality-specific temporal observations and a daily tabular feature vector. Formally,
\begin{equation}
\mathcal{X}^{\mathrm{ts}}
=
\left\{
\mathbf{X}_{ik}^{(m)}
\right\}_{m=1}^{M},
\qquad
\mathcal{X}^{\mathrm{tab}}
=
\left\{
\mathbf{z}_{ik}
\right\},
\end{equation}
where
$\mathbf{X}_{ik}^{(m)}
\in
\mathbb{R}^{T_{ik}^{(m)}\times S_m}$
denotes the observations of modality $m$ collected for subject $i$ on day $k$, with $T_{ik}^{(m)}$ observations and $S_m$ measurement channels. Since sensing modalities differ in their sampling characteristics, both $T_{ik}^{(m)}$ and $S_m$ vary across modalities. The vector
$\mathbf{z}_{ik}\in\mathbb{R}^{F}$
denotes the corresponding daily feature representation.

Raw observations were preprocessed independently for each modality before feature extraction. Stream-specific transformations were applied for numerical encoding, after which observations were restricted to valid monitoring days, duplicated timestamps were removed, and measurement channels were converted to \texttt{float32}.

\subsection{Feature Preparation}

\textbf{Feature Engineering. }
Daily features were extracted according to the characteristics of each modality:
\begin{itemize}
    \item $W_m$ and $P_c$: Daily channel-wise statistical descriptors were extracted. 
    \begin{itemize}
        \item For a univariate signal, features are calculated as:
            \begin{equation}
            \mathbf{f}(X)=
            \left[
            \operatorname{mean}(X),
            \operatorname{min}(X),
            \operatorname{max}(X),
            \operatorname{skew}(X),
            \operatorname{std}(X)
            \right].
            \end{equation}
        \item For multichannel signals, features were computed independently for each channel and concatenated.
        
    \end{itemize}

    \item $P_e$: Daily event counts were extracted.

    \item $W_s$: Daily summary measurements were used directly.

    \item $S_d$: The last daily response was retained. Questionnaire scores were normalized to $[0,1]$, whereas MoodLog ratings were linearly rescaled from $[-2,2]$ to $[0,1]$.
\end{itemize}
The resulting modality-specific feature vectors were concatenated to form the daily representation $\mathbf{z}_{ik}$.
\\ \\ 
\textbf{Feature Preprocessing. } 
Extracted features were preprocessed using feature-wise mean imputation and min--max scaling. Given  $F$ is the number of extracted features, feature dimensionality was subsequently reduced by univariate feature selection using ANOVA F-statistics, retaining the top-$K$ features:
\begin{equation}
K=\min\!\left(\max\!\left(20,\left\lfloor0.5F\right\rfloor\right),F\right),
\end{equation}

\subsection{Benchmark Design}
\textbf{Benchmark Configurations.}
For evaluation, the two smartphone modalities, event-based sensing ($P_e$) and continuous sensing ($P_c$), are merged into a unified smartphone feature group ($P$), while the weekly self-report modality ($S_w$) is excluded because its temporal resolution is not aligned with the remaining modalities.

Benchmark configurations are constructed from the feature space $\mathcal{M}$ to evaluate the effect of modality integration. Specifically, we consider predefined single-modality, two-modality, three-modality, and four-modality feature combinations, resulting in 15  configurations evaluated consistently across all baseline models and classification tasks. 

Let \(P=P_e\cup P_c\), the benchmark configuration space consists of four feature groups as follows:
\begin{equation}    
\mathcal{M}=\{W_m,\;W_s,\;S_d,\;P\},
\end{equation}

We exhaustively evaluate all possible non-empty combinations of these feature groups, yielding 15 benchmark configurations. These configurations are grouped according to the number of feature groups included:
\begin{itemize}
    \item \textbf{Single-group (4):} $W_m$, $W_s$, $S_d$, and $P$.
    \item \textbf{Two-group (6):} $W_m+W_s$, $W_m+S_d$, $W_m+P$, $W_s+S_d$, $W_s+P$, and $S_d+P$.
    \item \textbf{Three-group (4):} $W_m+W_s+S_d$, $W_m+W_s+P$, $W_m+S_d+P$, and $W_s+S_d+P$.
    \item \textbf{All-group (1):} $W_m+W_s+S_d+P$.
\end{itemize}

\textbf{Benchmark Tasks.}
To evaluate the utility of the proposed dataset under different clinical screening scenarios, we define four binary classification tasks as follows:

\begin{itemize}
\item \textbf{Task B1}: Classification between \textit{HC} and \textit{Dep}
\item \textbf{Task B2}: Classification between \textit{HC} and \textit{Anx}
\item \textbf{Task B3}: Classification between \textit{Dep} and \textit{Anx}
\item \textbf{Task B4}: Classification between \textit{HC} and clinical groups (\textit{Dep}+\textit{Anx}+\textit{OPC})
\end{itemize}

All benchmark tasks are evaluated using the same feature configurations, baseline models, and subject-wise cross-validation protocol to ensure fair comparison.

Table~\ref{tab:task_statistics} summarizes the number of subjects, samples, and class distributions across the four benchmark tasks. The benchmark tasks vary in sample size and exhibit moderate class imbalance, with the largest imbalance in Task B4.

\begin{table}[h]
\centering
\caption{Dataset statistics for the benchmark tasks. Class distributions are reported as class 0 / class 1.}
\label{tab:task_statistics}
\begin{tabular}{l r r c}
\toprule
Task & Subjects & Samples & Class Distribution \\
\midrule
\textbf{Task B1}: \textit{HC} vs.\ \textit{Dep} & 74 & 1,275 & 430 / 664 \\
\textbf{Task B2}: \textit{HC} vs.\ \textit{Anx} & 49 & 844 & 430 / 267 \\
\textbf{Task B3}: \textit{Dep} vs.\ \textit{Anx} & 61 & 1,057 & 664 / 267 \\
\textbf{Task B4}: \textit{HC} vs.\ \textit{Clinical} & 100 & 1,721 & 430 / 1,049 \\
\bottomrule
\end{tabular}
\end{table}

\newpage
\section{Experiment}
\label{sec:experiment}
This section presents the benchmark evaluation of the proposed dataset. We first describe the experimental setup, including the baseline models, feature configurations, and evaluation protocol, followed by quantitative benchmark results and analyses across different classification tasks.


\begin{table}[!h]
\centering
\small
\caption{Summary of benchmark evaluation settings.}
\label{tab:benchmark_setup}

\begin{tabular}{p{4cm}p{8cm}}
\toprule
\textbf{Component} & \textbf{Configuration} \\
\midrule

Cross-validation 
& 5-fold subject-wise Group K-Fold \\

Data split unit
& Participant level \\

Benchmark dimensions
& Tasks $\times$ Models $\times$ Feature groups \\

Feature settings
& Single- and multi-group configurations \\

Basline models
& LR, RF, XGB, MLP \\

Evaluation metric
& Subject-level F1-score \\

Value reports
& Mean $\pm$ standard deviation \\

\bottomrule
\end{tabular}

\end{table}


\begin{table}[!h]
\centering
\small
\caption{Hyperparameter settings of the baseline models.}
\label{tab:model_hyperparameters}

\begin{tabular}{p{4cm}p{8cm}}
\toprule
\textbf{Model} &
\textbf{Hyperparameters} \\
\midrule

LR &
Logistic Regression, $C=1.0$, default L2 regularization,
maximum iterations=1000, random state=42 \\

RF &
Random Forest, default hyperparameters,
random state=42 \\

XGB &
XGBoost, learning rate=0.05, number of estimators=300,
evaluation metric=logloss, random state=42 \\

MLP &
Multi-Layer Perceptron, default architecture,
maximum iterations=500, random state=42 \\

\bottomrule
\end{tabular}

\end{table}

\subsection{Experimental Setup}
\label{sec:experimental_setup}

\textbf{Baseline Models.}
We benchmark four machine learning models: Logistic Regression (LR), Random Forest (RF), XGBoost (XGB), and a Multi-Layer Perceptron (MLP). All models are trained using the same feature representations for each benchmark condition without model-specific feature engineering. Hyperparameter settings are summarized in Table~\ref{tab:model_hyperparameters}. 
Baseline models were implemented using the scikit-learn and the XGBoost library. Unless otherwise specified, default hyperparameter settings were used; the XGBoost classifier was configured with a learning rate of 0.05 and 300 estimators. A fixed random seed (42) was used to ensure reproducibility.

\textbf{Evaluation Metrics.}
We use subject-level macro F1-score (\texttt{mF1}) as the primary evaluation metric:

\[
mF_1 = \frac{2PR}{P+R},
\]

where $P$ and $R$ denote precision and recall, respectively. Predictions are evaluated at the subject level, and performance is reported as the mean $\pm$ standard deviation across five subject-wise cross-validation folds. F1-score is used to account for potential class imbalance in the classification tasks.

\textbf{Evaluation Protocol.}
All experiments are conducted using five-fold subject-wise Group K-Fold cross-validation, where samples from the same participant are restricted to a single fold to prevent subject-level information leakage. The same fold partition is used across all models, feature configurations, and classification tasks to ensure fair comparison. 

Each defined benchmark configuration defined is independently evaluated using all baseline models under the same subject-wise cross-validation protocol. This results in a complete benchmark matrix spanning all classification tasks, feature configurations, and baseline models.

All experiments are implemented in Python, using NVIDIA GeForce RTX 5080 GPU.

\newpage
\subsection{Baseline Results}
\textbf{Quantitative Analysis. }
Table~\ref{tab:benchmark_best} compares the performance of different feature combinations using the best-performing classifier for each binary classification task. Across the benchmark tasks, feature group combinations generally achieved higher performance than single-group configurations, although the optimal feature configuration varied across modeling tasks.
Comprehensive benchmark results for all evaluated feature combinations are provided in Appendix~A.
\begin{table}[!h]
\centering
\small
\begin{threeparttable}

\caption{Best benchmark performance across all classification tasks.}
\label{tab:benchmark_best}

\setlength{\tabcolsep}{6pt}
\renewcommand{\arraystretch}{1.0}

\begin{tabular*}{\textwidth}{@{\extracolsep{\fill}}lccc}
\toprule
\textbf{Modeling Tasks} &
\textbf{Best Configuration} &
\textbf{Best Model} &
\textbf{mF1 ($\uparrow$)} \\
\midrule

\textbf{Task B1}: \textit{HC} vs.\ \textit{Dep}
& $W_m + W_s + S_d$
& MLP
& \textbf{0.71} \scriptsize $\pm$0.17 \\

\textbf{Task B2}: \textit{HC} vs.\ \textit{Anx}
& $W_m + W_s + P + S_d$
& LR
& \textbf{0.69} \scriptsize $\pm$0.17 \\

\textbf{Task B3}: \textit{Dep} vs.\ \textit{Anx}
& $W_m$
& XGB
& \textbf{0.56} \scriptsize $\pm$0.22 \\

\textbf{Task B4}: \textit{HC} vs.\ (\textit{Dep+Anx+OPC})
& $W_m + W_s + S_d$
& LR
& \textbf{0.71} \scriptsize $\pm$0.06 \\

\bottomrule
\end{tabular*}

\begin{tablenotes}[flushleft]
\footnotesize
\item Results are reported as Subject-F1 (mean $\pm$ standard deviation) over five subject-wise cross-validation folds.
\end{tablenotes}

\end{threeparttable}
\end{table}

\begin{figure}[!h]
    \centering
    \includegraphics[width=1\columnwidth,keepaspectratio]{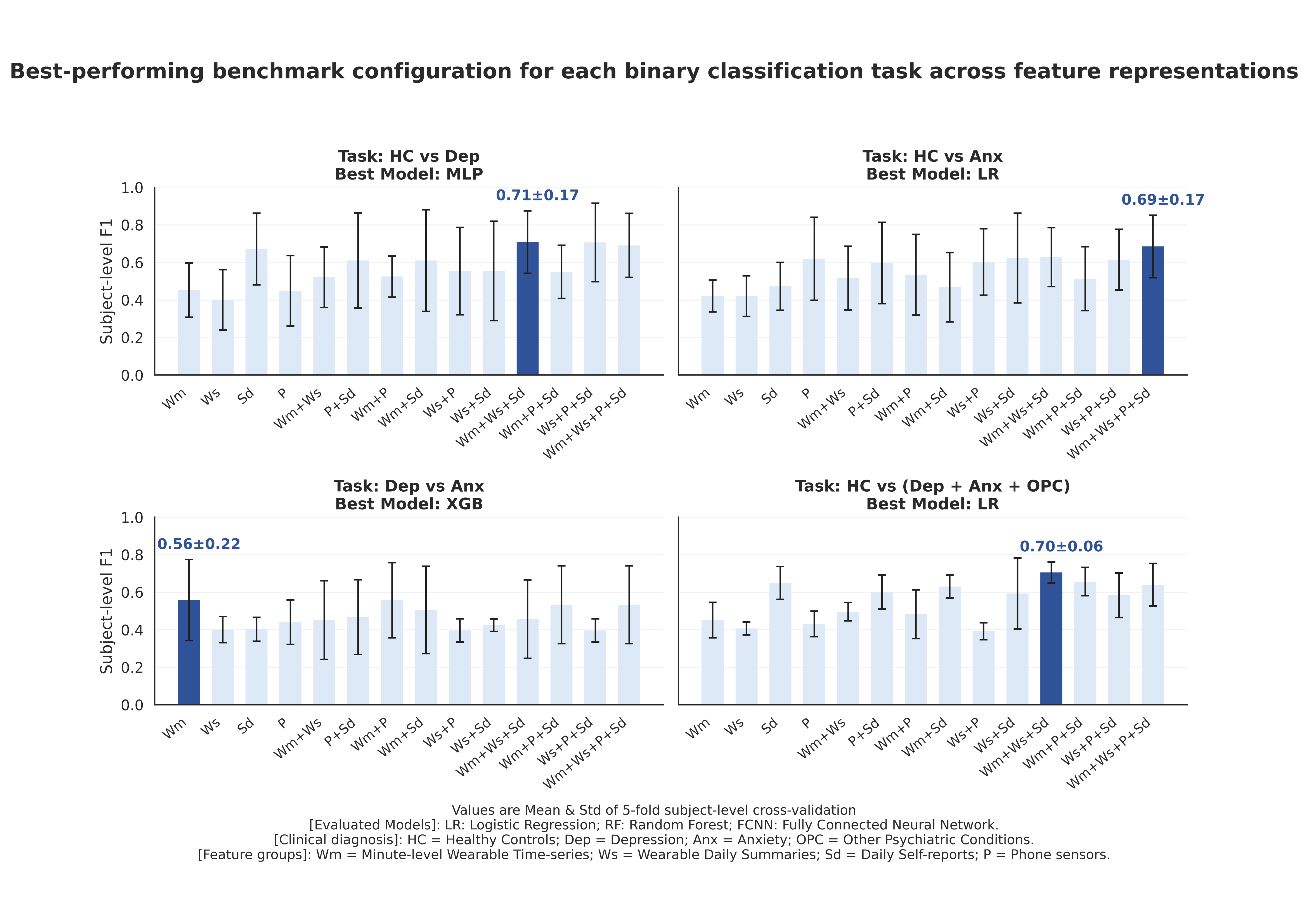}
    \caption{Performance comparison of different feature combinations for each binary classification task using the best-performing classifier. Bars show the mean Subject-F1, with error bars indicating the standard deviation over five subject-wise cross-validation folds. The highlighted bar denotes the best-performing feature combination for each task.}
    \label{fig:benchmark_barplot}
\end{figure}

Figure~\ref{fig:benchmark_barplot} summarizes benchmark performance across feature configurations grouped by the number of integrated feature modalities. Overall, configurations incorporating multiple feature groups tended to achieve higher performance, although the benefit of additional modalities varied across diagnostic tasks and feature combinations.
\newpage
\textbf{Effect of Multimodal Feature Integration. }
Table~\ref{tab:modality_summary} summarizes benchmark performance as a function of the number of integrated feature modalities. Overall, the average Subject-F1 increased progressively from single-modality to multimodal feature combinations, with the highest mean performance achieved when all four modalities were integrated. In contrast, the highest individual result was obtained using a three-modality configuration, suggesting that the optimal feature combination remained task-dependent.
\begin{table}[!h]
\centering
\small
\begin{threeparttable}

\caption{Performance grouped by feature group combination}

\label{tab:modality_summary}

\setlength{\tabcolsep}{18pt}
\renewcommand{\arraystretch}{1.0}

\begin{tabular}{lcc}
\toprule
\textbf{Benchmark Configuration} &
\textbf{mF1 ($\uparrow$)} &
\textbf{Range} \\
\midrule

Single-group
& 0.48 {\scriptsize $\pm$0.08}
& 0.35--0.67 \\

Two-group
& 0.52 {\scriptsize $\pm$0.08}
& 0.39--0.68 \\

Three-group
& 0.57 {\scriptsize $\pm$0.08}
& 0.40--0.71 \\

All-group
& \textbf{0.60} {\scriptsize $\pm$0.08}
& 0.43--0.69 \\

\bottomrule
\end{tabular}

\begin{tablenotes}[flushleft]
\footnotesize
\item Values are mean $\pm$ SD of subject-level Macro-F1 across configurations. Range indicates minimum--maximum Macro-F1.
\end{tablenotes}

\end{threeparttable}
\end{table}

\begin{figure}[!h]
    \centering
    \includegraphics[width=0.9\columnwidth,keepaspectratio]{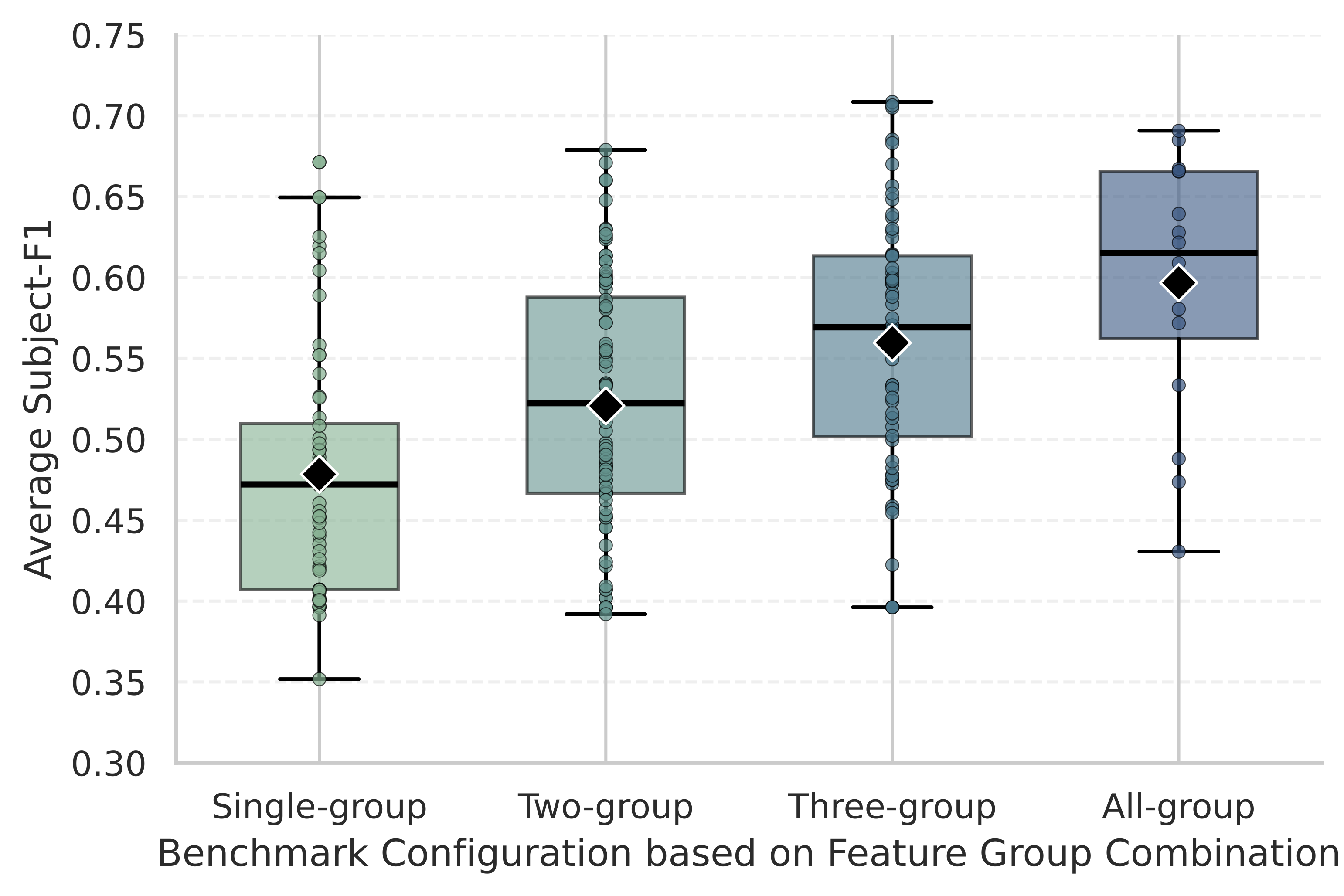}
    \caption{Distribution of subject-level Macro-F1 scores across benchmark configurations grouped by the number of feature groups. Boxplots show single-, two-, three-, and all-group combinations.}
    \label{fig:benchmark_boxplot}
\end{figure}

Figure~\ref{fig:benchmark_boxplot} summarizes the distribution of benchmark performance grouped by the number of feature modalities. Overall, configurations integrating a larger number of modalities tended to achieve higher Subject-F1 scores than single-modality configurations. The median performance increased progressively from one to four modalities, while the performance variance also became larger for multimodal configurations, reflecting greater diversity among feature combinations.

\newpage
\section*{Data Availability}
The Neurai-VN dataset \cite{cuong_q_pham_2026_21969065} is publicly available at \url{https://zenodo.org/records/18976768}.

\section*{Code Availability}
The code used to reproduce the benchmark experiments, including preprocessing pipelines and evaluation protocols, is publicly available at \url{https://github.com/neurai-vn/Neurai-VN-benchmark}.


\section*{Ethics and Privacy Statement}
The dataset used in this study was collected under ethical approval and informed consent, and all data were de-identified prior to analysis. This paper does not introduce any new data collection and solely reuses the existing dataset for computational evaluation.

\bibliographystyle{unsrt} 
\bibliography{refs}

\newpage
\appendix
\section{Appendix: Benchmark Results}
\label{app:benchmark}

\begin{table}[!h]
\centering
\small
\begin{threeparttable}

\caption{Performance comparison across different configurations for the HC vs.\ Dep classification task.}
\label{tab:hc_dep}
\setlength{\tabcolsep}{7pt}
\renewcommand{\arraystretch}{1.0}

\begin{tabular*}{\textwidth}{@{\extracolsep{\fill}}lcccc}
\toprule
\textbf{Feature Configurations} &
\textbf{LR} &
\textbf{RF} &
\textbf{XGB} &
\textbf{MLP} \\

\midrule
\multicolumn{5}{c}{\textit{Task B1: HC vs Dep}}\\
\midrule

\multicolumn{5}{l}{\textbf{Single-group}}\\
$W_m$ & 0.49 {\scriptsize $\pm$0.16} & 0.49 {\scriptsize $\pm$0.12} & 0.50 {\scriptsize $\pm$0.04} & 0.45 {\scriptsize $\pm$0.14} \\
$W_s$ & 0.42 {\scriptsize $\pm$0.20} & 0.35 {\scriptsize $\pm$0.08} & 0.51 {\scriptsize $\pm$0.27} & 0.40 {\scriptsize $\pm$0.16} \\
$S_d$ & 0.67 {\scriptsize $\pm$0.19} & 0.60 {\scriptsize $\pm$0.22} & 0.63 {\scriptsize $\pm$0.11} & 0.67 {\scriptsize $\pm$0.19} \\
$P$ & 0.39 {\scriptsize $\pm$0.13} & 0.48 {\scriptsize $\pm$0.15} & 0.49 {\scriptsize $\pm$0.17} & 0.45 {\scriptsize $\pm$0.19} \\
\midrule
\multicolumn{5}{l}{\textbf{Two-group}}\\
$W_m + W_s$ & 0.49 {\scriptsize $\pm$0.07} & 0.46 {\scriptsize $\pm$0.16} & 0.48 {\scriptsize $\pm$0.11} & 0.52 {\scriptsize $\pm$0.16} \\
$W_m + P$ & 0.49 {\scriptsize $\pm$0.08} & 0.49 {\scriptsize $\pm$0.11} & 0.52 {\scriptsize $\pm$0.10} & 0.52 {\scriptsize $\pm$0.11} \\
$W_m + S_d$ & 0.60 {\scriptsize $\pm$0.25} & 0.57 {\scriptsize $\pm$0.18} & 0.60 {\scriptsize $\pm$0.05} & 0.61 {\scriptsize $\pm$0.27} \\
$W_s + P$ & 0.41 {\scriptsize $\pm$0.13} & 0.46 {\scriptsize $\pm$0.15} & 0.48 {\scriptsize $\pm$0.12} & 0.55 {\scriptsize $\pm$0.23} \\
$W_s + S_d$ & 0.65 {\scriptsize $\pm$0.19} & 0.53 {\scriptsize $\pm$0.22} & 0.60 {\scriptsize $\pm$0.17} & 0.55 {\scriptsize $\pm$0.26} \\
$P + S_d$ & 0.63 {\scriptsize $\pm$0.18} & 0.60 {\scriptsize $\pm$0.18} & \textbf{0.63} {\scriptsize $\pm$0.23} & 0.61 {\scriptsize $\pm$0.25} \\
\midrule
\multicolumn{5}{l}{\textbf{Three-group}}\\
$W_m + W_s + S_d$ & 0.65 {\scriptsize $\pm$0.16} & \textbf{0.61} {\scriptsize $\pm$0.22} & 0.56 {\scriptsize $\pm$0.17} & \textbf{0.71} {\scriptsize $\pm$0.17} \\
$W_m + W_s + P$ & 0.53 {\scriptsize $\pm$0.03} & 0.49 {\scriptsize $\pm$0.16} & 0.57 {\scriptsize $\pm$0.13} & 0.56 {\scriptsize $\pm$0.13} \\
$W_m + P + S_d$ & 0.61 {\scriptsize $\pm$0.22} & 0.60 {\scriptsize $\pm$0.11} & 0.57 {\scriptsize $\pm$0.06} & 0.55 {\scriptsize $\pm$0.14} \\
$W_s + P + S_d$ & 0.60 {\scriptsize $\pm$0.23} & 0.60 {\scriptsize $\pm$0.26} & 0.61 {\scriptsize $\pm$0.28} & 0.71 {\scriptsize $\pm$0.21} \\
\midrule
\multicolumn{5}{l}{\textbf{All-group}}\\
$W_m + W_s + P + S_d$ & \textbf{0.67} {\scriptsize $\pm$0.18} & 0.60 {\scriptsize $\pm$0.24} & 0.58 {\scriptsize $\pm$0.14} & 0.69 {\scriptsize $\pm$0.17} \\

\bottomrule
\end{tabular*}
\begin{tablenotes}[flushleft]
\footnotesize
\item Results are reported as subject-level Macro-F1 ($\uparrow$), presented as mean $\pm$ standard deviation over five subject-wise cross-validation folds. The best mean score for each model is shown in \textbf{bold}.
\end{tablenotes}

\end{threeparttable}
\end{table}

\begin{table}[!h]
\centering
\small
\begin{threeparttable}
\caption{Performance comparison across different configurations for HC vs Anx classification task.
}
\label{tab:hc_dep}
\setlength{\tabcolsep}{7pt}
\renewcommand{\arraystretch}{1.0}

\begin{tabular*}{\textwidth}{@{\extracolsep{\fill}}lcccc}
\toprule
\textbf{Feature Configurations} &
\textbf{LR} &
\textbf{RF} &
\textbf{XGB} &
\textbf{MLP} \\
\midrule

\multicolumn{5}{c}{\textit{Task B2: HC vs Anx}}\\
\midrule

\multicolumn{5}{l}{\textbf{Single-group}}\\
$W_m$ & 0.42 {\scriptsize $\pm$0.08} & 0.48 {\scriptsize $\pm$0.08} & 0.47 {\scriptsize $\pm$0.14} & 0.54 {\scriptsize $\pm$0.21} \\
$W_s$ & 0.42 {\scriptsize $\pm$0.11} & 0.41 {\scriptsize $\pm$0.07} & 0.49 {\scriptsize $\pm$0.24} & 0.44 {\scriptsize $\pm$0.08} \\
$S_d$ & 0.47 {\scriptsize $\pm$0.13} & 0.47 {\scriptsize $\pm$0.13} & 0.47 {\scriptsize $\pm$0.13} & 0.47 {\scriptsize $\pm$0.13} \\
$P$ & 0.62 {\scriptsize $\pm$0.22} & 0.53 {\scriptsize $\pm$0.21} & 0.55 {\scriptsize $\pm$0.25} & 0.55 {\scriptsize $\pm$0.25} \\
\midrule
\multicolumn{5}{l}{\textbf{Two-group}}\\
$W_m + W_s$ & 0.52 {\scriptsize $\pm$0.17} & 0.49 {\scriptsize $\pm$0.08} & 0.54 {\scriptsize $\pm$0.18} & 0.61 {\scriptsize $\pm$0.23} \\
$W_m + P$ & 0.53 {\scriptsize $\pm$0.22} & 0.53 {\scriptsize $\pm$0.24} & 0.55 {\scriptsize $\pm$0.18} & 0.53 {\scriptsize $\pm$0.19} \\
$W_m + S_d$ & 0.47 {\scriptsize $\pm$0.18} & 0.53 {\scriptsize $\pm$0.12} & 0.47 {\scriptsize $\pm$0.08} & 0.47 {\scriptsize $\pm$0.11} \\
$W_s + P$ & 0.60 {\scriptsize $\pm$0.18} & 0.53 {\scriptsize $\pm$0.22} & 0.56 {\scriptsize $\pm$0.26} & 0.57 {\scriptsize $\pm$0.19} \\
$W_s + S_d$ & 0.62 {\scriptsize $\pm$0.24} & 0.45 {\scriptsize $\pm$0.13} & 0.49 {\scriptsize $\pm$0.19} & 0.51 {\scriptsize $\pm$0.18} \\
$P + S_d$ & 0.60 {\scriptsize $\pm$0.22} & \textbf{0.60} {\scriptsize $\pm$0.22} & 0.60 {\scriptsize $\pm$0.22} & \textbf{0.67} {\scriptsize $\pm$0.17} \\
\midrule
\multicolumn{5}{l}{\textbf{Three-group}}\\
$W_m + W_s + S_d$ & 0.63 {\scriptsize $\pm$0.16} & 0.60 {\scriptsize $\pm$0.19} & 0.60 {\scriptsize $\pm$0.17} & 0.65 {\scriptsize $\pm$0.24} \\
$W_m + W_s + P$ & 0.64 {\scriptsize $\pm$0.19} & 0.48 {\scriptsize $\pm$0.18} & 0.57 {\scriptsize $\pm$0.19} & 0.48 {\scriptsize $\pm$0.15} \\
$W_m + P + S_d$ & 0.51 {\scriptsize $\pm$0.17} & 0.56 {\scriptsize $\pm$0.22} & 0.62 {\scriptsize $\pm$0.14} & 0.55 {\scriptsize $\pm$0.21} \\
$W_s + P + S_d$ & 0.61 {\scriptsize $\pm$0.16} & 0.57 {\scriptsize $\pm$0.19} & 0.53 {\scriptsize $\pm$0.15} & 0.60 {\scriptsize $\pm$0.13} \\
\midrule
\multicolumn{5}{l}{\textbf{All-group}}\\
$W_m + W_s + P + S_d$ & \textbf{0.69} {\scriptsize $\pm$0.17} & 0.57 {\scriptsize $\pm$0.21} & \textbf{0.67} {\scriptsize $\pm$0.18} & 0.61 {\scriptsize $\pm$0.17} \\

\bottomrule
\end{tabular*}
\begin{tablenotes}[flushleft]
\footnotesize
\item Results are reported as subject-level Macro-F1 ($\uparrow$), presented as mean $\pm$ standard deviation over five subject-wise cross-validation folds. The best mean score for each model is shown in \textbf{bold}.
\end{tablenotes}

\end{threeparttable}
\end{table}

\newpage
\begin{table}[!t]
\centering
\small
\begin{threeparttable}
\caption{Performance comparison across different configurations for Dep vs Anx classification task.
}
\label{tab:hc_dep}
\setlength{\tabcolsep}{7pt}

\begin{tabular*}{\textwidth}{@{\extracolsep{\fill}}lcccc}
\toprule
\textbf{Feature Configurations} &
\textbf{LR} &
\textbf{RF} &
\textbf{XGB} &
\textbf{MLP} \\
\midrule

\multicolumn{5}{c}{\textit{Task B3: Dep vs Anx}}\\
\midrule

\multicolumn{5}{l}{\textbf{Single-group}}\\
$W_m$ & 0.41 {\scriptsize $\pm$0.07} & 0.51 {\scriptsize $\pm$0.09} & \textbf{0.56} {\scriptsize $\pm$0.22} & 0.48 {\scriptsize $\pm$0.14} \\
$W_s$ & 0.41 {\scriptsize $\pm$0.07} & 0.40 {\scriptsize $\pm$0.07} & 0.40 {\scriptsize $\pm$0.07} & 0.40 {\scriptsize $\pm$0.07} \\
$S_d$ & 0.40 {\scriptsize $\pm$0.06} & 0.40 {\scriptsize $\pm$0.06} & 0.40 {\scriptsize $\pm$0.06} & 0.41 {\scriptsize $\pm$0.07} \\
$P$ & 0.40 {\scriptsize $\pm$0.06} & 0.40 {\scriptsize $\pm$0.06} & 0.44 {\scriptsize $\pm$0.12} & 0.46 {\scriptsize $\pm$0.20} \\
\midrule
\multicolumn{5}{l}{\textbf{Two-group}}\\
$W_m + W_s$ & 0.41 {\scriptsize $\pm$0.07} & 0.47 {\scriptsize $\pm$0.07} & 0.45 {\scriptsize $\pm$0.21} & 0.42 {\scriptsize $\pm$0.08} \\
$W_m + P$ & 0.40 {\scriptsize $\pm$0.06} & \textbf{0.52} {\scriptsize $\pm$0.08} & 0.56 {\scriptsize $\pm$0.20} & 0.45 {\scriptsize $\pm$0.15} \\
$W_m + S_d$ & 0.40 {\scriptsize $\pm$0.06} & 0.47 {\scriptsize $\pm$0.07} & 0.51 {\scriptsize $\pm$0.23} & 0.45 {\scriptsize $\pm$0.15} \\
$W_s + P$ & 0.47 {\scriptsize $\pm$0.20} & 0.40 {\scriptsize $\pm$0.06} & 0.40 {\scriptsize $\pm$0.06} & 0.45 {\scriptsize $\pm$0.15} \\
$W_s + S_d$ & 0.40 {\scriptsize $\pm$0.06} & 0.40 {\scriptsize $\pm$0.07} & 0.42 {\scriptsize $\pm$0.03} & 0.41 {\scriptsize $\pm$0.07} \\
$P + S_d$ & 0.40 {\scriptsize $\pm$0.06} & 0.40 {\scriptsize $\pm$0.06} & 0.47 {\scriptsize $\pm$0.20} & 0.50 {\scriptsize $\pm$0.22} \\
\midrule
\multicolumn{5}{l}{\textbf{Three-group}}\\
$W_m + W_s + S_d$ & \textbf{0.48} {\scriptsize $\pm$0.20} & 0.47 {\scriptsize $\pm$0.07} & 0.46 {\scriptsize $\pm$0.21} & 0.42 {\scriptsize $\pm$0.13} \\
$W_m + W_s + P$ & 0.40 {\scriptsize $\pm$0.06} & 0.47 {\scriptsize $\pm$0.07} & 0.53 {\scriptsize $\pm$0.21} & 0.46 {\scriptsize $\pm$0.09} \\
$W_m + P + S_d$ & 0.40 {\scriptsize $\pm$0.06} & 0.52 {\scriptsize $\pm$0.08} & 0.53 {\scriptsize $\pm$0.21} & \textbf{0.51} {\scriptsize $\pm$0.23} \\
$W_s + P + S_d$ & 0.47 {\scriptsize $\pm$0.20} & 0.40 {\scriptsize $\pm$0.06} & 0.40 {\scriptsize $\pm$0.06} & 0.50 {\scriptsize $\pm$0.19} \\
\midrule
\multicolumn{5}{l}{\textbf{All-group}}\\
$W_m + W_s + P + S_d$ & 0.43 {\scriptsize $\pm$0.02} & 0.49 {\scriptsize $\pm$0.10} & 0.53 {\scriptsize $\pm$0.21} & 0.47 {\scriptsize $\pm$0.12} \\

\bottomrule
\end{tabular*}
\begin{tablenotes}[flushleft]
\footnotesize
\item Results are reported as subject-level Macro-F1 ($\uparrow$), presented as mean $\pm$ standard deviation over five subject-wise cross-validation folds. The best mean score for each model is shown in \textbf{bold}.
\end{tablenotes}

\end{threeparttable}
\end{table}

\begin{table}[!h]
\centering
\small
\begin{threeparttable}
\caption{Performance comparison across different configurations for HC vs (Dep+Anx+OPC) classification task.
}
\label{tab:hc_dep}
\setlength{\tabcolsep}{7pt}

\begin{tabular*}{\textwidth}{@{\extracolsep{\fill}}lcccc}
\toprule
\textbf{Feature Configurations} &
\textbf{LR} &
\textbf{RF} &
\textbf{XGB} &
\textbf{MLP} \\
\midrule

\multicolumn{5}{c}{\textit{Task B4: HC vs (Dep+Anx+OPC)}}\\
\midrule

\multicolumn{5}{l}{\textbf{Single-group}}\\
$W_m$ & 0.45 {\scriptsize $\pm$0.09} & 0.45 {\scriptsize $\pm$0.08} & 0.47 {\scriptsize $\pm$0.13} & 0.49 {\scriptsize $\pm$0.10} \\
$W_s$ & 0.41 {\scriptsize $\pm$0.03} & 0.43 {\scriptsize $\pm$0.03} & 0.40 {\scriptsize $\pm$0.03} & 0.44 {\scriptsize $\pm$0.12} \\
$S_d$ & 0.65 {\scriptsize $\pm$0.09} & 0.62 {\scriptsize $\pm$0.12} & 0.59 {\scriptsize $\pm$0.11} & 0.65 {\scriptsize $\pm$0.09} \\
$P$ & 0.43 {\scriptsize $\pm$0.07} & 0.46 {\scriptsize $\pm$0.09} & 0.53 {\scriptsize $\pm$0.11} & 0.50 {\scriptsize $\pm$0.08} \\
\midrule
\multicolumn{5}{l}{\textbf{Two-group}}\\
$W_m + W_s$ & 0.50 {\scriptsize $\pm$0.05} & 0.43 {\scriptsize $\pm$0.07} & 0.47 {\scriptsize $\pm$0.11} & 0.52 {\scriptsize $\pm$0.17} \\
$W_m + P$ & 0.48 {\scriptsize $\pm$0.13} & 0.58 {\scriptsize $\pm$0.14} & 0.53 {\scriptsize $\pm$0.11} & 0.68 {\scriptsize $\pm$0.13} \\
$W_m + S_d$ & 0.63 {\scriptsize $\pm$0.06} & 0.53 {\scriptsize $\pm$0.10} & 0.55 {\scriptsize $\pm$0.14} & 0.61 {\scriptsize $\pm$0.11} \\
$W_s + P$ & 0.39 {\scriptsize $\pm$0.04} & 0.48 {\scriptsize $\pm$0.09} & 0.48 {\scriptsize $\pm$0.08} & 0.51 {\scriptsize $\pm$0.14} \\
$W_s + S_d$ & 0.59 {\scriptsize $\pm$0.19} & 0.58 {\scriptsize $\pm$0.18} & 0.58 {\scriptsize $\pm$0.12} & 0.59 {\scriptsize $\pm$0.09} \\
$P + S_d$ & 0.60 {\scriptsize $\pm$0.09} & 0.66 {\scriptsize $\pm$0.09} & 0.63 {\scriptsize $\pm$0.09} & 0.66 {\scriptsize $\pm$0.14} \\
\midrule
\multicolumn{5}{l}{\textbf{Three-group}}\\
$W_m + W_s + S_d$ & \textbf{0.70} {\scriptsize $\pm$0.06} & 0.52 {\scriptsize $\pm$0.11} & 0.50 {\scriptsize $\pm$0.12} & 0.60 {\scriptsize $\pm$0.05} \\
$W_m + W_s + P$ & 0.45 {\scriptsize $\pm$0.07} & 0.57 {\scriptsize $\pm$0.13} & 0.59 {\scriptsize $\pm$0.11} & \textbf{0.69} {\scriptsize $\pm$0.16} \\
$W_m + P + S_d$ & 0.66 {\scriptsize $\pm$0.08} & 0.60 {\scriptsize $\pm$0.05} & 0.59 {\scriptsize $\pm$0.18} & 0.63 {\scriptsize $\pm$0.16} \\
$W_s + P + S_d$ & 0.58 {\scriptsize $\pm$0.12} & \textbf{0.67} {\scriptsize $\pm$0.11} & \textbf{0.64} {\scriptsize $\pm$0.08} & 0.68 {\scriptsize $\pm$0.10} \\
\midrule
\multicolumn{5}{l}{\textbf{All-group}}\\
$W_m + W_s + P + S_d$ & 0.64 {\scriptsize $\pm$0.11} & 0.67 {\scriptsize $\pm$0.09} & 0.62 {\scriptsize $\pm$0.08} & 0.63 {\scriptsize $\pm$0.11} \\

\bottomrule
\end{tabular*}
\begin{tablenotes}[flushleft]
\footnotesize
\item Results are reported as subject-level Macro-F1 ($\uparrow$), presented as mean $\pm$ standard deviation over five subject-wise cross-validation folds. The best mean score for each model is shown in \textbf{bold}.
\end{tablenotes}

\end{threeparttable}
\end{table}

\end{document}